\documentclass[11pt,a4paper,twocolumn]{article}
\usepackage{arxiv}
\usepackage{times,latexsym}
\usepackage{url}
\usepackage{amsmath}
\usepackage{bbm}
\usepackage{diagbox}
\usepackage{graphicx}
\usepackage[T1]{fontenc}
\usepackage{fixltx2e}
\usepackage{pifont}

\usepackage[square,sort,comma,numbers]{natbib}
\usepackage{doi}
\usepackage{soul}
\usepackage{subcaption}

\usepackage{graphicx} %
\usepackage[table]{xcolor}
\usepackage{url}
\usepackage{physics}
\usepackage{tikz-cd}
\usepackage{hyperref}

\newcommand{\secref}[1]{Section~\ref{sec:#1}}

\newcommand{\figref}[1]{Figure~\ref{fig:#1}}
\newcommand{\figstworef}[2]{Figures~\ref{fig:#1} and~\ref{fig:#2}}
\newcommand{\tabref}[1]{Table~\ref{tab:#1}}

\DeclareRobustCommand{\DE}[3]{#2}

\usepackage[framemethod=TikZ]{mdframed}
\usepackage{xcolor}
\usepackage{listings}
\usepackage{caption}
\newmdenv[%
    backgroundcolor=blue!5,
    linecolor=black,
    outerlinewidth=0.01pt,
    font=\footnotesize,
]{promptbox}
\makeatletter
\let\orig@promptbox=\promptbox
\def\promptbox{
  \@ifnextchar[{\promptbox@opt}{\orig@promptbox}
}
\def\promptbox@opt[#1]{
  \orig@promptbox[frametitle={#1}]
}
\makeatother

\title{Will GPT-4 Run DOOM?}
\renewcommand{\headeright}{}

\usepackage{etoolbox}
\makeatletter
\pretocmd\@maketitle
  {\def\@makefnmark{\hbox{\@textsuperscript{\normalfont\@thefnmark}}}}
  {}{\FAIL}
\makeatother
\usepackage{authblk}

\author[1]{%
{Adrian de Wynter\thanks{The University of York. adrian.dewynter@york.ac.uk}%
}
}
\begin{document}
\maketitle

\begin{abstract}
We show that GPT-4's reasoning and planning capabilities extend to the 1993 first-person shooter Doom. 
This large language model (LLM) is able to run and play the game with only a few instructions, plus a textual description--generated by the model itself from screenshots--about the state of the game being observed. 
We find that GPT-4 can play the game to a passable degree: it is able to manipulate doors, combat enemies, and perform pathing. 
More complex prompting strategies involving multiple model calls provide better results. 
While further work is required to enable the LLM to play the game as well as its classical, reinforcement learning-based counterparts, we note that GPT-4 required no training, leaning instead on its own reasoning and observational capabilities. 
We hope our work pushes the boundaries on intelligent, LLM-based agents in video games. 
We conclude by discussing the ethical implications of our work. %

\end{abstract}

\section{Introduction}

Large language models (LLMs) such as GPT-4 \cite{GPT4} are able to parse input text (prompts) as instructions, and execute them to remarkable accuracy. 
No training or fine-tuning is required for this: instead, a few exemplars (shots) in the prompt suffice. 
This capability has led the community to actively research LLM problem-solving capabilities,  
Of interest to us are planning and reasoning skills in complex environments, namely, video games. 

In this paper, we aim to push and evaluate these capabilities by playing a 1993 first-person shooter (FPS) developed by id Software: Doom.\footnote{All products, company names, brand names, trademarks, and images are
 properties of their respective owners.} 
Although it has become a running meme to determine whether a device is able to run a port of its code (``will it run Doom?''), this exploration has scientific value. 
For starters, the source code is open-source; and has been ported into Python, a common scripting language in machine learning research. 
Doom is also a relatively simple, non-deterministic game, so LLM memorisation \cite{PlagiariseLee,carlini2022quantifying,dewynter2023evaluation} of benchmarks is not a concern. 
Hence, the game can serve as an informal, community-led benchmark on computational capabilities, and, as proposed here, on LLM planning abilities.

We show that, although GPT-4 has technical limitations on its ability to run Doom directly, it can still act as a proxy for the engine--not unlike some existing ``will it run (display) Doom?'' implementations, such as Windows Notepad \cite{NotepadDoom}, \emph{E. Coli} \cite{GutDoom}, or Minecraft \cite{MinecraftDoom}. 
We also demonstrate that using two GPT-4 models, the game can be played by the LLM to a reasonable degree: %
it is capable of open doors, fight enemies, and fire weapons. 
It is also capable of executing a broader instruction set (a level walkthrough) to improve its own performance. 
A critical evaluation of its reasoning capabilities shows that the model is able to execute short-term planning, but it has shallow reasoning depth and low memory recall: for example, if an enemy goes out of view, the model will forget about it.

This paper adds to the growing body of work informing the application and development of LLM-based intelligent agents in video games, for example, to develop playtesters \cite{IntelligentAgents} that do not require pretraining. 
To our knowledge, our work is the first to deploy LLMs in FPS games, and the first to use visual input from the game instead of descriptions. 
It is also the first to measure the performance of LLMs in single-player FPS. 
We open source the code used on this work.\footnote{Code with licencing restrictions, videos of the model playing the game, and accessible screenshots, are available at \url{https://adewynter.github.io/Doom}.}

\section{Related Work}

The focus of our paper is on LLM-based intelligent agents in video games. 
However, the idea of employing machine learning to play video games, and Doom in particular, is not new. 
For example, Alvernaz and Togelius \cite{8080408} used autoencoders for simple pathing under a sandbox environment; and Lample and Chaplot \cite{FPSDoom} employed reinforcement learning (RL) to outmatch humans in various FPS, Doom included. 
Further work applied and refined deep RL approaches by providing environmental signals \cite{FPSHierarchicalRL}; and by including previous history \cite{DeepRLHistory}. Extensions to other genres \cite{CardRTS} and multiple agents \cite{DeepQFactorised} also have been proposed.

\subsection{Reasoning and Planning in LLMs}

It is well-known that the models are very sensitive to the prompt's phrasing and exemplars \cite{liu-etal-2022-makes,lu-etal-2022-fantastically}. 
Although chain-of-thought approaches are robust to this pathology \cite{ dewynter2023i,ChainOfThought,NEURIPS2022_8bb0d291}, research has shown that this approach fails in inductive reasoning tasks \cite{humanlike,Dziri2023FaithAF}, especially under fictional worlds \cite{saparov2023language}, symbolic reasoning \cite{dewynter2023i}, and deeper reasoning paths \cite{Dziri2023FaithAF,anil2022exploring}. 

Since generalised planning involves an arbitrary reasoning depth, there have been efforts to benchmark LLM capabilities around formal reasoning and planning \cite{suzgun-etal-2023-challenging,anonymous2023put,huang2022language}. 
Pallagani et al. \cite{PlanningSurvey} evaluated the performance of LLMs in planning, and generalised planning in particular. 
They found that the models evaluated were acceptable, but not effective, at generalised planning. 

That said, carefully-designed approaches such as PEARL \cite{sun2023pearl}, RAP \cite{RAP}, and k-level reasoning \cite{klevels} have shown to improve over the standard reasoning approaches such as zero-shot and chain-of-thought, when baselined against GPT-4, as well as other LLMs \cite{planandsolve,LLMPLanner}. 
Moreover, it is known that adding multimodal knowledge to problem solving improved performance in spatial reasoning \cite{liu-etal-2022-things}, and even text-only tasks \cite{lu-etal-2022-imagination,yang-etal-2022-z,10.1145/3462244.3479965}, and planning \cite{anonymous2023multimodal,wang-etal-2023-multimedia}.

\subsection{LLMs in video games}

LLMs have been used with varying degrees of success to play video games accross various genres, such as Minecraft \cite{wang2023describe}, Pok\'{e}mon Battles \cite{pokellmon}, and Starcraft II \cite{starcraftiillms}. 
The last two have been proposed as benchmarks for evaluating the LLM's ability to play video games \cite{starcraftiillms,pokellmon}, due to the finite nature of the game: each session is a single match with a clear winner. 
To our knowledge, our work is the first to deploy LLMs to work in FPS, and the first to utilise images from the video game as opposed to textual descriptions. It is also the first to measure the performance of LLMs in single-player FPS.

\section{Background and Setup}

\subsection{GPT-4}

GPT-4 is a LLM developed by OpenAI with several variants. 
Relevant to this work are the GPT-4 with Vision (or GPT-4V) and the text-to-text (GPT-4) variants.\footnote{\url{https://platform.openai.com/docs/guides/vision} and \url{https://platform.openai.com/docs/guides/text-generation/faq}, accessed 30 January 2024.} 

GPT-4V accepts images and textual input, and returns text. 
Input images may be up to 20MB, with the option of a behind-the-scenes resizing step. 
All our images are 640x400, the original resolution of Doom. 
We left the resizing as automated, yet noticed no drop in quality from the model's output.

The version of GPT-4 we used accepts and returns text, with a maximum combined input-output limit of 32K tokens. 
This large token limit was desirable so we could fit a short history of the model's actions in previous turns. 
The authors claim that the reasoning capabilities of this model are state-of-the-art \cite{GPT4}. 

No details have been released around architecture, training algorithms, or data. 
We, however, assume that this model has not been trained to play Doom.

\subsection{Doom}

Doom is a 2.5D\footnote{The map appears 3D, but the objects in-game are rendered as 2D, and so is the traversable environment.} FPS mostly developed in C. 
The player must traverse a level fighting enemies and avoiding environmental hazards. 
Power ups (health, armour, ammo, invulnerability, etc.) and additional weapons are scattered through the map. 
The player UI has health, armour, and ammo counters. See \figref{VisionSample} for an in-game screenshot.

Since the game is 2.5D, there is no aiming on the y-axis (i.e., looking up/down) and all is needed from the user is to align the centre of the screen to the enemy or item to be acted upon. 
Its low complexity in terms of playability, portability, and graphics makes it an ideal candidate for our evaluation. 

From a computational perspective, Doom falls into the (large) category of games in PSPACE \cite{ALOUPIS2015135}. 
That is, given enough memory--polynomial on the size of the game state--a Turing machine could solve (play) this game. 
Given that the engine running the game must be able to handle the game itself, this suggests a loose bound on the ability of an automated player to solve it. 
In the case of GPT-4, the memory available (token limit) is equivalent to the description of about 16 640x400 screenshots. Hence, it cannot retain sufficient information to simulate the full game, and it must rely on heuristics. 

\subsection{Experimental Setup}

Our experimental setup consists of two main components: a \textbf{Vision} component (calling GPT-4V) that takes in screenshots from the running engine, and returns structured descriptions of the game state; and an \textbf{Agent} model (calling GPT-4) that makes decisions based on Vision's output and the previous history. 
Vision and Agent's prompts has been tuned to return parseable outputs that can be used with the engine. In particular, Agent's outputs are only from actions taken from to the allowed set of keystrokes. 

For the more complex prompts, we also use a GPT-4 model to perform either finer-level planning (\textbf{Planner}) or provide expert advice (\textbf{Experts}). 
The Planner is a different prompt designed to take in the current state and the walkthrough, and output a finer-grained, step-by-step natural language plan. 
We opted against using common planning languages, such as PDDL, to better lean on GPT-4's linguistic capabilities. 
The Experts in our setup are simply extra, separate calls to a different version of Agent.

\begin{figure}
    \centering
    \includegraphics[width=\columnwidth]{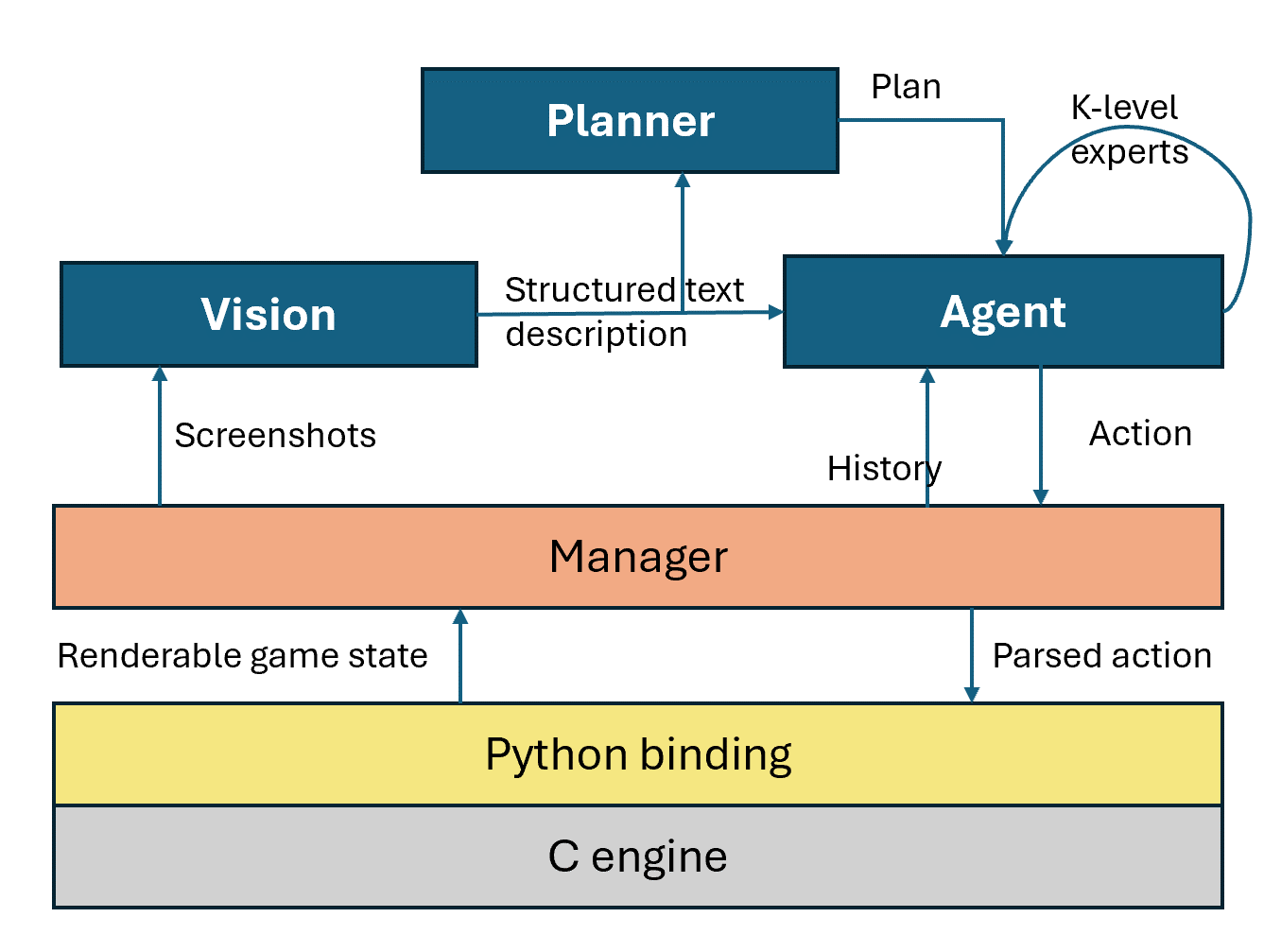}
    \caption{Architecture of our system. The core piece of our code, the Manager, is a Matplotlib interface able to run Doom on top of the Python binding, itself an interface for the original Doom code in C. 
    The Manager communicates states and actions to the binding; sends screenshots to Vision, and retains and parses the history of previous moves by Agent, Planner and Experts (if applicable).
    The three boxes above describe our prompt setup: Vision feeds descriptions to Planner and Agent, and an extra set of calls is performed to obtain Experts.}
    \label{fig:setup}
\end{figure}

\begin{figure}
    \centering
    \includegraphics[width=\columnwidth]{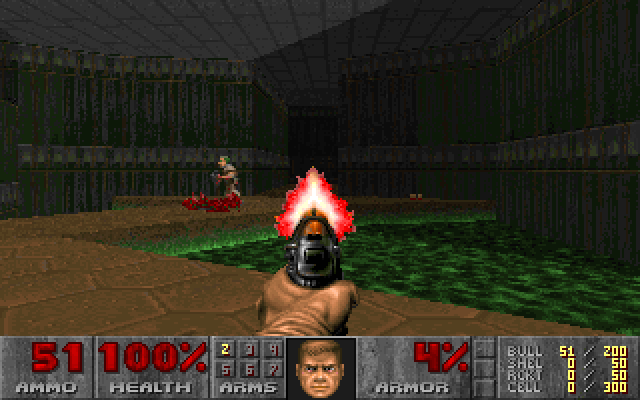}
    \caption{Sample screenshot fed into Vision, taken directly from room C in E1M1. 
    The output from the model is in \figref{visionoutput}.}
    \label{fig:VisionSample}
\end{figure}

See \figref{VisionSample} for a sample screenshot and Vision model output, and \figref{setup} for a diagram of our setup including Planner and Experts. 
All model calls are done through the Azure Open AI API. 

The underlying engine is an open-source Python binding to the C Doom engine\footnote{\url{https://github.com/wojciech-graj/cyDoomgeneric}} running on Matplotlib. 
We modify this binding to allow for calls to Vision and Agent, as well as to skip the start and quit sequences of commands. 
It is worth noting that, in earlier tests, Agent was able to select a level and difficulty in the startup screen, but for the sake of consistency on level and difficulty setting we hard-coded the commands. 

\subsection{Prompts}

For all levels we tested a \textbf{na\"{i}ve} prompt that solely took the previous output history and output the current action and reasons as to why that output was chosen, analogous to Wang et al. \cite{wang2023describe}; and three planning prompts, all variations on the na\"{i}ve prompt:
\begin{enumerate}
    \item A \textbf{walkthrough} prompt that takes a broad, step-by-step description of the level and what to do (``Step 1: cross the room, kill the enemies, go to the next room'', e.g.). 
    Following Huang et al. \cite{huang2022language}, we separate the walkthrough in steps for easier parsing and planning. 
    \item A \textbf{plan}-generating prompt, where an extra component (the Planner) takes in the walkthrough and the history; and outputs a more granular plan to be executed by the walkthrough prompt.\footnote{Also known as \emph{hierarchical planning}, see, e.g., Wang et al. \cite{planandsolve} and Song et al. \cite{LLMPLanner} for LLM applications.}
    \item A \textbf{k-levels} prompt \cite{klevels}, which behaves like the plan prompt, but also takes in the moves of three Experts (all calls to the plan prompt) and outputs its own move based on this input. 
    This prompt style has been shown to improve reasoning in dynamic contexts. In this paper we only do $k=2$. 
\end{enumerate}

All prompts are symbolic chain-of-thought, which was shown to provide good results in complex scenarios with multiple referents and long-term dependencies \cite{dewynter2023i}. 
This type of prompting requires the model to fill out a set of variables (as opposed to reasoning steps) in a multiturn setting before making its decision. 
We also include a request for an explanation, as it is known to increase performance in out-of-distribution observations \cite{Brahman_Shwartz_Rudinger_Choi_2021}. 
Sample prompts and the walkthrough are available in the Appendix. 
Ahead of the experiments we tuned the prompts to ensure consistent performance. 

\subsection{Optimisations}

We reduced the frame rate by half (i.e., we only call the model every two frames) similar to other approaches \cite{FPSDoom}. This has the advantage to save a call in the case of pressed keys. 
For example, if Agent says ``LEFT'', the input to the engine would be a sequence of actions analogous to pressing and depressing the left arrow key. 
For motion actions, we ``press'' the key 3 times before releasing it, and once otherwise. 
We tested the frame rate with and without halving and found no noticeable changes in performance. 
The game itself ran at an average of $8$ frames per second under normal (in-engine, Python binding, non-GPT-4) play. 

GPT-4 is relatively reliable when following the prompt's output format and desired output set. 
However, parsing errors do occur. To mitigate this, we introduced an extra action (``WAIT'') to skip that pair of frames' desired actions. ``WAIT'' is within the allowable output set of the model, but also the default action whenever Agent's output is not within that set. 

Ahead of time we tuned the temperature and token limits and settled on temperatures of $0.1$ and $0.9$; and maximum tokens of $2880$ and $25$ for Vision and Agent, respectively. 
High randomness and a low token output was desirable for Agent, but the opposite for Vision. 
We retain the same parameters for Agent when calling each Expert, but the Planner was set at temperature $0.1$ and $150$ tokens. 
To improve throughput, Planner and Expert calls occur only once every $30$ and $60$ frames, respectively. 

For our design of the walkthrough, we skip all shortcuts and secrets.

\begin{figure}
\centering
\begin{promptbox}
\begin{lstlisting}
In the screenshot, we see a first-person view of a player
holding a pistol mid-fire, with a small muzzle flash
visible. The player appears to be in a room with green
floors and brown walls that have a metallic texture. There
is a pool of green liquid, which could be toxic, on the
floor. In the distance, there is a red-colored enemy,
which resembles a demon, and it appears to be either
injured or in an attack animation, as there is a splash of
red pixels around its midsection, possibly indicating
blood or damage.

The HUD (Heads-Up Display) at the bottom of the
screen shows various information:

- On the left, there is an ammo count displaying "51" for
the current weapon.
- In the center, there is a health percentage indicator
showing "100%
- To the right of the health indicator, there are four
grayed-out weapon icons, suggesting that those weapons
are not currently in the player's possession.
- Further to the right, there is an armor percentage
indicator showing "4%
- At the very bottom right, there are numerical counts
for different types of ammunition: "BULL" (bullets) at
50, "SHEL" (shells) at 0, "ROCK" (rockets) at 0, and
"CELL" (energy cells) at 200.
\end{lstlisting}
\end{promptbox}
\caption{Vision output corresponding to the screenshot from \figref{VisionSample}. 
The model was requested to return a description of the room and the UI, with the UI as a list. This structured output allowed us to maintain consistency between calls. Remark that the model was not informed about either the hazards of the pools, or the meaning of the ammo counters. The prompt is in the Appendix.}
\label{fig:visionoutput}
\end{figure}

\subsection{Play Settings}\label{sec:playsettings}

We evaluated the first map (``Hangar'') of the first episode (``Knee Deep in the Dead''); under the same difficulty (``Hurt Me Plenty''; the third, out of five, difficulty ranks). This map is referred to as E1M1, and can be seen in \figref{e1m1} along with a typical human traversal without secrets. 

The walkthrough we wrote (in the Appendix) instructs the model to traverse sections of the map in sequence: the first room (A in \figref{e1m1}); the computer room (r. B); the zig-zag bridge room (r. C); and the end sequence (r. D). 
It ignores secret areas and indicates the location of enemies and environmental hazards. 

We ran each prompt ten times on the same level, and left the model to play the game uninterrupted. 
We only intervened (terminated the game) when the model had been stuck for more than 1,000 frames on the same location. 

\begin{figure}
    \centering
    \includegraphics[width=\columnwidth]{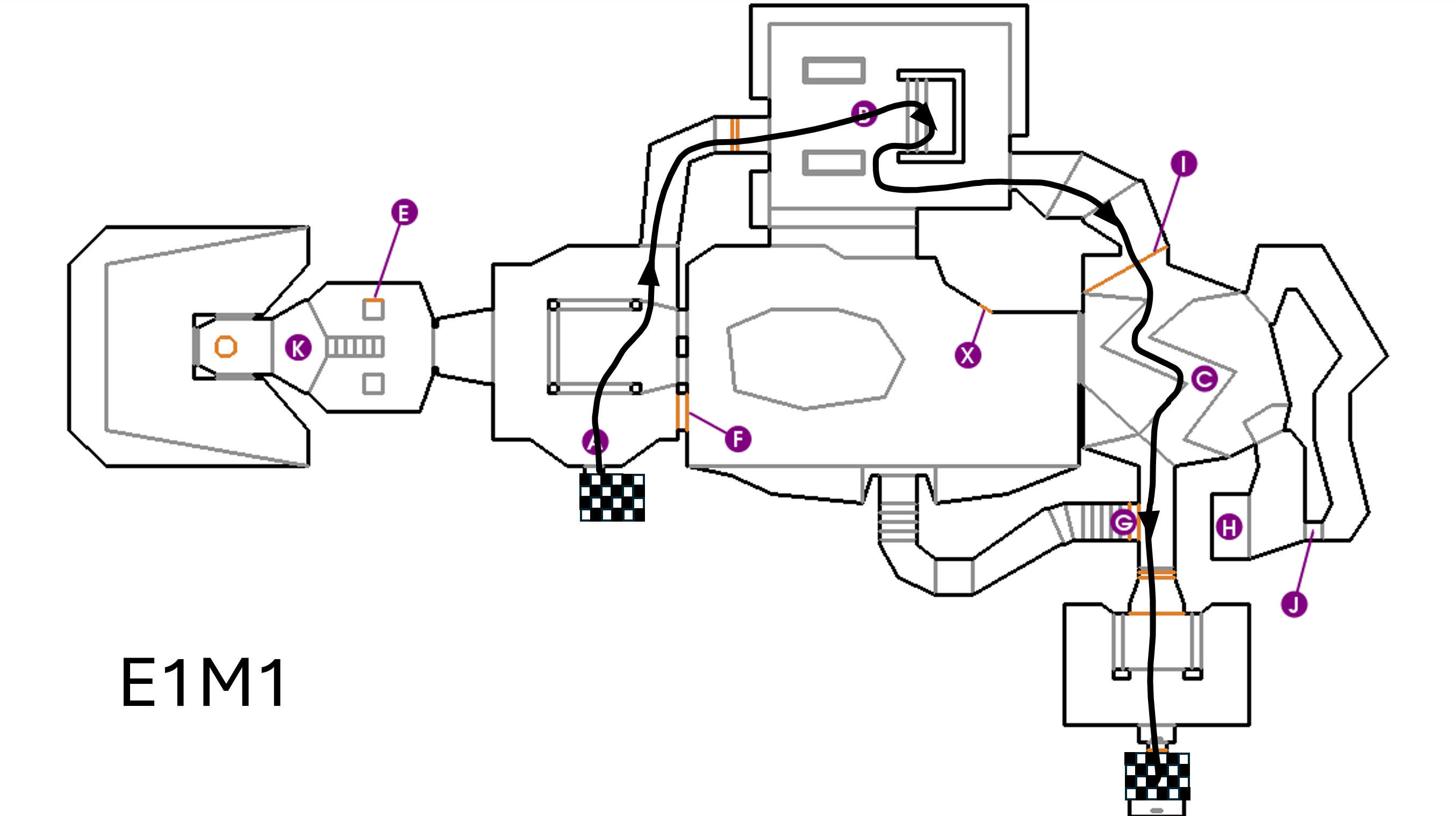}
    \caption{Map evaluated in this paper, corresponding to Episode 1, Map 1 (E1M1, ``Hangar'') of Doom \cite{DoomWiki}. The trajectory from our walkthrough follows the sequence A, B, C, D; starting at A (leftmost checkered flag) and ending in D (rightmost checkered flag). 
    Depicted is a typical human run with no exploration. 
    The player must flip a switch in D to end the game. 
    Doors are denoted by an orange line. 
    The zig-zag room (C) contains pools of acid (not pictured) that damage the player if walked on.}
    \label{fig:e1m1}
\end{figure}

\subsection{Metrics}

In a single-player FPS involving traversing a level, metrics are somewhat imprecise: a match can span multiple savefiles and deaths, and, arguably, the winning condition is to successfully finish the game (or level); and it may not be necessary to kill all enemies in the level. 

We modify slightly the Passed Maps' Average Time (PMAT) metric by Song \textit{et al.} \cite{FPSHierarchicalRL}. 
Similar to PMAT, we measured the time (in frames) that a player took to play the game, but we segment it by the frames spent in each of the rooms (A, B, C, D in \figref{e1m1}). 
We also separately measure PMAT by weighing deaths, and call this metric \emph{Death-Weighted} PMAT, or D-PMAT. 

Namely, suppose there is a set of trials (runs) $T = \{t_1, t_2, \dots t_n\}$, where $t_i$ is a collection of time-sorted frames corresponding to a specific trial. 
Let $T[p:q] \subset T$ be the subset of time-continuous frames spent in a stage delimited by sections $p$ and $q$. 
Then:

\begin{align}
    \text{PMAT}_{p:q} &= \frac{1}{\vert T[p:q] \vert}\sum_{t \in T[p:q]} \left\vert t \right\vert, \\
    \text{D-PMAT}_{p:q} &= \frac{1}{\vert T[p:q] \vert}\sum_{t \in T[p:q]} \left\vert t \right\vert + \mathbbm{1}_t(\text{death})\lambda,
\end{align}

where $\mathbbm{1}_t(\text{death})$ is an indicator function (zero if no deaths in trial $t$, one otherwise), and $\lambda$ is a constant, set in our case to $\lambda = 1,000$ (i.e., the timeout from \secref{playsettings}). 
For a segment $X$ that the model never visited in any $t \in T$, PMAT$_X$ is undefined ($\infty$). 
If the player backtracks, it is counted as the last room entered. 
We compare these to average human runs collected under the same setup (i.e., no shortcuts).

\section{Experiment Results}

\begin{figure*}[h]
    \centering
    \includegraphics[width=\textwidth]{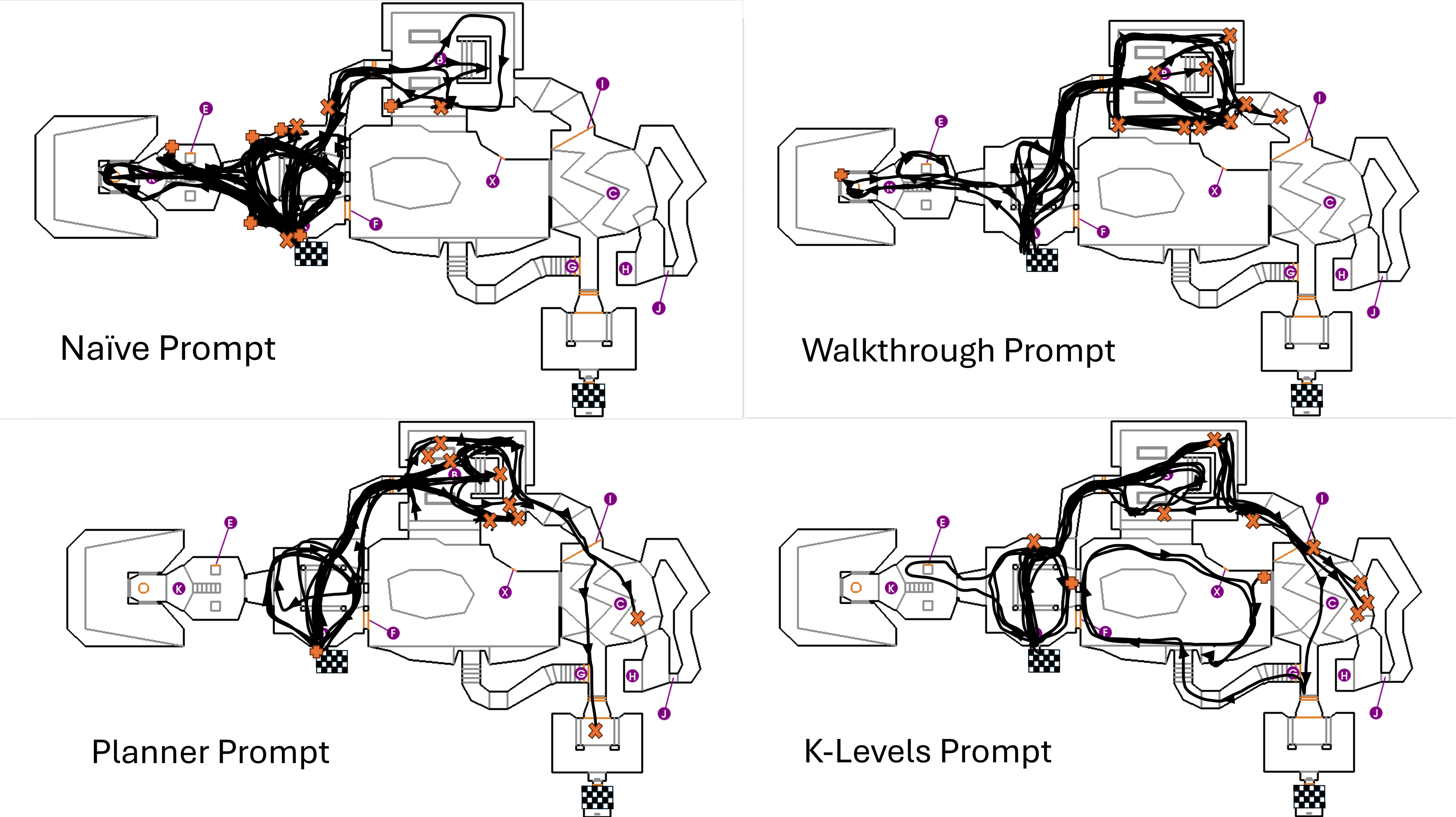}
    \caption{Trajectories by Agent. The starting point is at the leftmost checkered flag; the ending (a switch that must be activated) is the rightmost checkered flag. 
    Other areas (e.g., G, J) are secrets. 
    Time-outs are denoted by \ding{58} and deaths by \ding{54}. 
    \textit{Clockwise from the top left}: na\"ive, with walkthrough, with k-levels, and with planner. The na\"ive prompt could not leave the first room and got frequently stuck. 
    It died twice at the hand of the zombies in room B; once due to shooting an explosive barrel. 
    The walkthrough prompt made it to the second room, but got stuck often in the corners and was killed by the zombies. 
    The planning prompt was able to kill the zombies in room B, and almost finished the map.  
    K-levels was consistently able to reach room C, although died by stepping on the acid. 
    It once managed to open a secret room, but could not backtrack and timed out.}
    \label{fig:trajectories}
\end{figure*}

Our D-PMAT and PMAT results are shown in \tabref{results}. 
Given that our metrics only capture play time, we also record paths commonly followed by the model and show them in \figref{trajectories}. 
Every trial was manually examined to determine the frames where the model entered each area (delimited by the door threshold) and whether timeouts or deaths had occurred. 

\begin{center}
\begin{table*}[ht]
\centering
\begin{tabular}{| c || p{0.24\columnwidth} | p{0.24\columnwidth} |p{0.24\columnwidth} | p{0.24\columnwidth} || c | c | c |} \hline
\textbf{Prompt} & PMAT$_A$ / D-PMAT$_A$ & PMAT$_B$ / D-PMAT$_B$ & PMAT$_C$ / D-PMAT$_C$ & PMAT$_D$ / D-PMAT$_D$ & Deaths & Timeouts & Finish \\ \hline\hline
Na\"ive         & 1409/1509 & 634/1413          & $\infty$  & $\infty$ & $40\%$ & $60\%$ & No \\ \hline
Walkthrough     & 671/671 & 657/1212 & 259/1259 & $\infty$ & $90\%$ & $10\%$ &No \\ \hline
Plan         & 559/559 & 903/1570 & \cellcolor{blue!15}193/860  & \cellcolor{blue!15}47/1047 & $90\%$ & $10\%$ & No \\ \hline
K-Levels        & \cellcolor{blue!15}434/534 & \cellcolor{blue!15}671/796 & 731/1588 & $\infty$ & $80\%$ & $20\%$ & No \\ \hline\hline
Human           & 78/78 & 108/108 & 158/158 & 104/104& $0\%$  & $0\%$ & Yes \\ \hline 
\end{tabular}
\caption{PMAT and D-PMAT results per room for our prompts when compared with an average human player. Highlighted in blue are the best-performing prompts for each stage averaged over our metrics. 
Neither prompt was able to finish the game, although planning and k-levels were remarkably close. The na\"ive prompt was unable to consistently get out of room A and timed out frequently. 
The plan prompt improved performance over the walkthrough prompt, enabling Agent to traverse the map more effectively. 
K-levels was also able to get through the first rooms relatively fast, though typically dying when stepping on acid. 
We attribute the better performance of planning and k-levels to the prompt's ability to obtain fresher context, enabling Agent to recalculate paths. 
Note how the PMAT$_D$ for Planner is better than the human baseline, but the D-PMAT$_D$ correctly encodes that the model was unable to finish.}\label{tab:results}
\end{table*}
\end{center}

\subsection{Play Performance}
The model was unable to finish the map under any of the prompts. 
However, it came close twice: once in the plan prompt, entering room D and immediately dying at the hands of the imp awaiting there; and once with k-levels, where it got to the door of room D, backtracked, and opened a secret door leading to the courtyard, where it timed out. 
In all prompts, Agent performed motions and actions in line with the instructions. 
It could open doors (even when backtracking), return fire, and in more advanced versions of the prompts (plan and k-levels) it was even able to get unstuck from corners. K-levels was particularly effective at this. 

However, Agent frequently overlooked enemies, had poor aim, and ignored environmental hazards. 
In a few limited instances, it shot point blank walls, or worse, explosive barrels.

\subsection{Pathing Performance}

The model usually hugged the walls when traversing the map, facing them and walking at an angle. 
In some cases it got stuck behind barrels or corners, but after a few hundred frames it could re-plot a trajectory. 

We observed that pathing was noticeably better under the walkthrough-enabled prompts (walkthrough, plan, k-levels). 
The re-plotting mentioned earlier was noticeably better and faster with k-levels, as long as the call frequency was about three times that of Planner (60 frames versus 30). 
When the frequencies were the same, Experts would hallucinate and introduce too much noise in the call. Agent was not be able to get out of room A in this scenario.

\section{Discussion}

To understand the performance of our system we analysed the call logs, and, in particular, the explanations given by the model. 
We observed that GPT-4 generally output the correct actions in context, but the rationale behind these was poor and fraught with hallucinations: 
for example, Planner tended to indicate incorrect steps from the walkthrough; 
or the Experts justified an action based on a counterfactual statement (e.g., ``activating the switch requires moving UP''). 

This suggests that the model is not able to understand its position in the world, yet somehow be able to play the game to an acceptable degree, returning fire, dodging, and operating doors, as long as these actions did not require planning spanning more than a few frames. 
That is, it is capable of some level of contextual inference but not long-term planning or reasoning. 
Indeed, this also could explain why Agent tended to run in circles, overlooking enemies and eventually dying at their hands. 

The high variability in the trajectories may be attributed to the observations above, but also to prompt sensitivity--a well-known problem of LLMs \cite{lu-etal-2022-fantastically}--and to the high temperatures used on Experts and Agent. 

That said, walkthrough-enabled prompts were noticeably better at traversing the map, and the addition of exemplars to improve traversal (e.g. to avoid getting stuck in corners) was effective at forcing Agent to output new trajectories after a few frames. 
These prompts were also much more effective at fighting enemies, as long as the enemies were within sight. 
This suggests that the model's planning capabilities are enhanced by as many contextual clues as possible, in-line with similar findings around reasoning in tasks like argument mining \cite{dewynter2023i}. 
Further work could develop significantly more detailed prompts to explore this observation.

\subsection{Limitations}

\paragraph{Real-Time Play}
The Agent-Vision combination, let alone with the addition of Planner or Experts, is not ready for real-time play: each inference step (per frame) for the na\"ive prompt alone takes approximately a minute to run. 
Even with our optimisations, this process places it outside of the realm of practical applications. 

\paragraph{Metrics}
Our evaluation focused on the ability of the model to play the game, and found satisfactory results. 
However, it is likely that under other metrics, such as human-like play \cite{behaviouralcloning,navigateme}, the model would not fare well, especially due to its tendency to hug, and sometimes shoot, the walls. 
The prompt's planning efficiency (as opposed to success) could be measured using either Success weighted by Path Length (SPL) \cite{embodiedeval} or Distance to Success (DTS) \cite{semanticExploration}, both measured by map size. 

\paragraph{Reward Model}
All our approaches involved prompting and rewarding based on in-context learning. 
Approaches that require back-propagation, such as RAP, were not considered given the current feasibility of fine-tuning GPT-4. 
Smaller models that are more approachable could be a better fit for this task, given that it is known that fine-tuning does improve planning performance \cite{PlanningSurvey}. 

\paragraph{Prompting}
Our testing was limited to four different types of prompting strategies. Given the model's sensitivity to the prompt and the dynamicity of the environment, it is probable that other prompts could provide different results. 
Moreover, it is probable that exemplars taken from actual in-game performances of the model will yield better performance than our handcrafted exemplars. 
Our exploration also suggested that planning may be enhanced by multiple contextual clues. 
We leave these observations for further work. 

\section{Ethical Implications}
In this work, GPT-4 showed remarkable capabilities given its setup: 
with only a simple depiction of the world and its rules, it could operate doors, follow simple objectives, and identify enemies and fire weapons at them. 

Although LLMs have been applied for good (take, for example, automated rewriting for dyslexic users \cite{Lampost}); 
the relative ease at which this model was able to pick up the rules in a first-person shooter, added to the simplicity of the code and approaches, should at least raise some alarms around possible misuse. 
Indeed, this calls for a much more careful evaluation and regulation of this technology. 

The work presented here has technical limitations, as discussed on the previous section. 
However, with how rapid the field is progressing, we expect some, if not all, of these limitations to be lifted. 
To paraphrase \emph{Jurassic Park}, we as scientists should be more preoccupied now on whether we \emph{should}, as opposed to whether we \emph{could}. 

\section{Conclusion}

In this paper we evaluated whether GPT-4 was able to run Doom, and whether it was able to play the game. 
We found that, although the model was not entirely successful at finishing the map, it displayed remarkable agency throughout the game, with more complex prompting strategies delivering consistently better results. 
That said, this illusion of agency was lost somehow when looking at the rationale behind some of the model's decisions, which showcased a lack of understanding of the task. 
This suggests that this model's reasoning depth is shallow, as it is only reliable within a few frames. 

Further work could improve upon the planning portions of our work; or create standardised benchmarks for LLMs beyond using the single-player mode. 
This can be achieved by using open-source libraries such as VizDoom \cite{Wydmuch2019ViZDoom}. 
Also, it is known that fine-tuning improves LLM planning capabilities \cite{PlanningSurvey}. 
Although this is not currently possible for GPT-4--the focus of our work--other smaller models, such as Llama 2 \cite{llamav2} have come close to this LLM's capabilities and could be better candidates for fine-tuning-based strategies. 

One surprising finding of our paper was this model's level of agency, along with the ease of access and simplicity of the code. 
This suggests a high potential for misuse. 
We release the code to contribute to the development of better video game agents, but we call for a more thorough regulation effort for this technology.

\bibliographystyle{IEEEtran}
\DeclareRobustCommand{\DE}[3]{#3}
\bibliography{biblio}

\appendix
\section{Prompts}

In this section we show sample prompts used for our work (\figstworef{visionprompt}{naiveprompt}) and the walkthrough used in the planner, k-levels and walkthrough prompts (\figref{walkthroughdesc}).

\begin{figure}[h]
\centering
\begin{promptbox}
\begin{lstlisting}
This a screenshot from DOOM. Give me a description of
the screenshot showing what you see, and what is in the
HUD, in that order.
\end{lstlisting}
\end{promptbox}
\caption{Vision prompt. Although less complex than the prompts used for Agent, we noticed that tuning this prompt further was detrimental for downstream performance. 
Moreover, Vision consistently output the desired syntactical structure and features (e.g., ammo counts, health).}
\label{fig:visionprompt}
\end{figure}

\begin{figure}[h]
\centering
\begin{promptbox}
\begin{lstlisting}
1. You start out the game in a room with a blue pool.
2. Go across the pool.
3. Keep going straight into the hallway.
4.Go straight down the hall, and open the door. Enter the
computer room.
5. Kill the two guys there. Go around the room with blue
walls, and to the back right of the computer room, where
there is an entrance to the bridge room (if you face the 
grey door, you need to turn around).
6. Make a right, kill the two guys, and cross the bridge.
Don't fall into the green acid pool.
7. Go straight and open the door at the end of the bridge
room. Kill the imp behind the door. Enter the loot room.
8. Go straight the loot room and open the next door.
Enter the switch room.
9. Inside the switch room there is a switch on the right
wall. Hit the switch after getting the items, if you need 
them.
\end{lstlisting}
\end{promptbox}
\caption{Walkthrough for E1M1, ``Knee Deep in the Dead''. Note that the names are slightly different (bridge room as opposed to zig-zag room, e.g.).}
\label{fig:walkthroughdesc}
\end{figure}

\begin{figure}
\centering
\begin{promptbox}
\begin{lstlisting}
# Instructions:
You are a videogame playing bot. You are playing 1993 
DOOM.
Your job is to play the game: traverse the level, kill 
enemies, ensure you have ammo and health.
Every turn, take ONE action based on the state given; 
and the history of what you did before.
# Available actions:
- UP: move forward
- DOWN: move backwards
- LEFT: turn left
- RIGHT: turn right
- STRAFE LEFT: strafe left
- STRAFE RIGHT: strafe right
- FIRE: fire your equipped weapon
- USE: use the item
- WAIT: wait for one second
- SPEED: move faster or slower
- 1: select weapon 1 (if available)
- 2: select weapon 2 (if available)
- 3: select weapon 3 (if available)
- 4: select weapon 4 (if available)
- 5: select weapon 5 (if available)
- 6: select weapon 6 (if available)
- 7: select weapon 7 (if available)
# Additionally:
- If the History shows that you haven't left the room, try 
exploring (UP/DOWN/LEFT/RIGHT), using items or 
doors. Walk UP to the item to collect them. Walk UP t
doors, and USE it to open it.
- If the History shows that you have too many WAITs, 
UPs, DOWNs, LEFTs, or RIGHTs, try exploring 
(UP/DOWN/LEFT/RIGHT), using items or doors. Walk
UP to the item or door, and USE it.
- If your health starts to decrease, you are taking 
damage! Align your weapon with the enemy and shoot
it. If you can't see them, they are behind you! Turn
around! (LEFT, LEFT or RIGHT, RIGHT)
- Make sure to avoid walking on acid!
- You cannot jump or go through windows.
- If your HUD says that your health is 0%
output GAME OVER instead, but only if that is the case
- The player means YOU.

# Examples:
<EXEMPLARS GO HERE>

Play begins here!
|History|
<HISTORY>

State:
<VISION_RESPONSE>
|Action|
\end{lstlisting}
\end{promptbox}
\caption{Na\"ive prompt, with exemplars omitted for brevity. The instructions address common errors that Agent showed in preliminary tests, such as trying to go through windows and walking on acid. 
Other prompts involve more structured input (e.g. plans, k-level outputs) but are generally similar to this prompt to maintain consistency.}
\label{fig:naiveprompt}
\end{figure}

\end{document}